\documentclass[11pt]{article}

\usepackage[final]{acl}

\usepackage{times}
\usepackage{latexsym}
\usepackage{multirow,multicol}
\usepackage{adjustbox}
\usepackage{booktabs}
\usepackage{pgfplots}
\pgfplotsset{compat=1.18}
\usepackage[table]{xcolor}
\definecolor{darkgreen}{rgb}{0.0, 0.5, 0.0}
\usepackage{linguex}
\usepackage{subcaption}
\usepackage{tabu}

\usepackage[T1]{fontenc}

\usepackage[utf8]{inputenc}

\usepackage{microtype}

\usepackage{inconsolata}

\usepackage{graphicx}

\usepackage[most]{tcolorbox}
\usepackage{xcolor}

\definecolor{lightgray}{RGB}{240,240,240} 
\definecolor{ethosblue}{RGB}{0,102,204}
\definecolor{pathosorange}{RGB}{255,140,0}
\usetikzlibrary{patterns}

\title{How Ethos and Pathos Appeals Resonate in Reader Interpretations of Social Media Messages}

\author{
  \textbf{Ewelina Gajewska\textsuperscript{*}},
  \textbf{Katarzyna Budzynska\textsuperscript{*}},
  \textbf{Jarosław A. Chudziak\textsuperscript{*}},
  \textbf{Liesbeth Allein\textsuperscript{$\dagger$$\ddagger$}}
\\
\\
  \textsuperscript{*}Warsaw University of Technology, Poland \\
  \textsuperscript{$\dagger$}Department of Computer Science, KU Leuven, Belgium \\
  \textsuperscript{$\ddagger$} Department of Electronics and Information Systems, Ghent University, Belgium\\
  \small{\textbf{Correspondence}: \texttt{ewelina.gajewska.dokt@pw.edu.pl}} }

\begin{document}

\maketitle 
\begin{abstract}
Rhetorical strategies and their influence on audiences are often studied through social media posts and comments. However, this focus overlooks the universal audience, which is the majority of readers who remain silent and do not explicitly express how a message affects them. This study investigates how two classical modes of persuasion, ethos and pathos, resonate in the silent audience’s interpretations of meaning. Using a dataset of social media sentences paired with human-written interpretations, we label both sources for ethos and pathos and assess whether these rhetorical appeals are preserved. Our analyses show that interpretations diverge from the original sentences in 30\% of cases, with rhetorically charged content eliciting greater variability than neutral content. We further find that ethos and pathos in original sentences can predict audience attitudes toward the author, underscoring the subtle ways rhetoric shapes perception beyond visible engagement. 
\end{abstract}

\section{Introduction}

Rhetoric, meaning, and audience perception are deeply intertwined in online communication. Yet analyzing their relationship is particularly challenging in ambiguous contexts such as social media. Ambiguity creates an \textit{interpretive space} \citep{sandri-etal-2023-dont}, allowing the same message to be understood in multiple ways depending on the cultural background, knowledge, and cognitive frames of different audiences. 
Rhetorical devices are central to how this ambiguity is negotiated. This is clear in current digital ecosystems, where prominent figures exercise considerable rhetorical power. Through appeals to credibility (\textit{ethos}) and emotion (\textit{pathos}) \citep{Aristotle1991}, they shape how people perceive themselves, others, and society as a whole. {Unlike devices like metaphors or narratives, these appeals directly target interpersonal trust and affective orientation.} Investigating how such {foundational modes of persuasion} resonate with audiences is paramount to studying influence, trust, and polarization online \citep{del2016echo,garimella2018political,buder2021does}. 

So far, computational works on rhetoric mainly focus on identifying rhetorical strategies and fallacies 
in social media posts and comments \citep{habernal-etal-2017-argotario,da-san-martino-etal-2019-fine,sheng-etal-2021-nice,jin-etal-2022-logical,wiegand-etal-2023-euphemistic,chen-etal-2024-exploring-potential,bagdon-etal-2025-donate,ramponi-etal-2025-fine}. However, inferring audience perceptions only from explicit content presents strong limitations as only a small share of the audience is captured. 
The dynamics of interpretation and influence extend beyond vocal participants to the much larger \textit{universal audience}. {In classical and contemporary rhetorical theory \citep{Aristotle1991,perelman1971rhetorique}, they are those who do not speak but remain rhetorically present, i.e., they observe, internalize, and shape discourse. 
This concept aligns with social media dynamics, where a majority of users consume content without commenting. In that respect, this work studies the rhetorical situation of interpretation rather than demographic segmentation.}

\textbf{This study investigates \textit{if} and \textit{how} rhetorical devices transform as they move from the original sentence to audience interpretations.} {Specifically, we examine whether ethos and pathos appeals in the sentence are retained, transformed, or omitted in internal representations of sentence meaning, i.e., reader interpretations. Interpretations are strong predictors of persuasive influence \citep{eagly2014recipient}. While influence can occur without full reproduction of devices, what is retained or omitted reveals which cues are salient and memorable. We further explore whether these appeals are predictive for variation in reader attitudes towards the sentence's author and describe additional factors that guide interpretation alongside ethos and pathos labels.}


\paragraph{Contributions}

{Our work introduces \textit{a novel computational perspective on rhetorical influence} by examining how rhetorical devices resonate in audience interpretations of social media content. We provide \textit{large-scale empirical evidence} of ethos and pathos transfer from text to audience interpretations. This type of perspectivism has been rarely explored in computational research. Through quantitative and qualitative} analyses, we identify \textit{predictive factors driving ethos and pathos divergence} between source sentences and audience interpretations, and within interpretations. We further \textit{enrich an existing interpretation modeling dataset with new annotation layers of ethos and pathos}, which are made publicly available\footnote{\url{https://doi.org/10.17605/OSF.IO/DAC2T}}.

\section{Related Work} \label{related_work}

\paragraph{Variation in interpretation.} This work is inspired by a growing literature advocating perspectivist approaches to language understanding \citep{plank2022problem,cabitza2023toward,fleisig2024perspectivist,frenda2025perspectivist}. Perspectivism moves away from single ground-truth interpretations to capture human variation. It is motivated by strong subjectivity and frequent annotation disagreement in tasks such as natural language inference \citep{pavlick2019inherent,lee-etal-2023-large,weber-genzel-etal-2024-varierr}, toxic and offensive language detection \citep{goyal2022your,sandri-etal-2023-dont,davani2024disentangling,trusca-allein-2025-mimicking}, and values identification \citep{sorensen2024value}. 

Unlike dominant approaches in perspectivism literature, this work does not assess the subjectivity of ethos and pathos through human label variation of the original sentence. Instead, we capture variation in language understanding through the set of audience interpretations that accompany the original sentence and how these differ in appeals to ethos and pathos. Our analysis of ethos and pathos thus takes place after interpretations are formulated. Consequently, the translation from original sentence to reader interpretations is not explicitly constrained by prior judgments of ethos and pathos, which is arguably closer to how interpretation occurs ``in the wild''.

\paragraph{Ethos and pathos.}
Many computational studies on rhetoric focus on rhetorical fallacies \citep{habernal-etal-2018-name,goffredo2022fallacious,jin-etal-2022-logical,musi2022developing,glockner-etal-2025-grounding,ramponi2025fine}, which may function as persuasive shortcuts \citep{walton2010fallacies}. However, the influence of classic rhetorical strategies like pathos\footnote{Related tasks such as emotion detection and sentiment analysis have a longstanding tradition in natural language processing. Yet, while these tasks center on the identification of emotional content expressed in text, pathos mining investigates the strategic use of emotions for persuasion.} and, especially, ethos on the audience remains underinvestigated. \citet{evgrafova-etal-2024-analysing} analyzed pathos in user-generated argumentation and highlighted how emotional appeals can influence audience attitudes and intensify divisions. \citet{duthie-ethos} built a computational model for extracting ethotic expressions from the UK parliamentary debates.  \citet{gajewska2024digital} extended their approach to social media contexts, investigating how ethos attacks and supports contribute to the polarization of attitudes in climate change discussions.

This study investigates if and how ethos and pathos change when a sentence is translated to different interpretations among its audience.

\section{Methodology}

\subsection{Data}

Our analyses rely on the OrigamIM dataset \citep{allein2024origamim}, which was developed to support interpretation modeling \citep{allein2025interpretation}. It contains 2,018 English sentences sampled from subreddit r/ChangeMyView, which represent views on controversial and polarizing topics, such as abortion and racism. In this subreddit, users articulate their stance on a topic and invite others to persuade them to reconsider. Each sentence is paired with approximately five human-written reformulations (9,851 interpretations in total) that reflect readers' interpretations of its implicit meanings, guided by diverse annotations of hidden moral judgments. Each interpretation is also paired with an attitude score on a five-point Likert scale, ranging from \textit{very negative} (1) to \textit{very positive} (5), indicating the reader's attitude toward the author of the original sentence. 

The annotators in OrigamIM processed the text presented to them and formed internal interpretations, akin to a universal audience. Since they were unconstrained in phrasing, interpretations may rephrase, summarize, or even expand the sentence's content. Summarization often highlights salient cues, whereas expansion reveals interpretive ambiguity. Note that recording these interpretations served as a methodological step to capture cognitive responses, not evidence of active engagement.

Our analyses require multiple human-written interpretations and a context where rhetorical strategies are actively employed. OrigamIM satisfies these conditions, making it an ideal setting for isolating rhetorical effects. Platforms like X or Facebook differ in interaction style (e.g., more reactive) and may exhibit persuasion less prominently, which could reduce precision for the type of analysis conducted here. While five interpretations cannot fully capture the diversity of the universal audience \citep{allein2025interpretation}, this indicative sample of human-written audience interpretation provides a solid starting point for systematic study of rhetorical resonance.

\subsection{Ethos and Pathos Classification}
For this preliminary study, we opt for a silver standard annotation pipeline to classify the rhetorical dimensions. Manually annotating a dataset of this scale, comprising nearly 10,000 text spans, for complex rhetorical dimensions like \textit{ethos} and \textit{pathos} presents significant resource and time constraints that fall outside the scope of this initial work. 
Furthermore, there exist datasets for training automated classifiers on these specific rhetorical appeals, therefore leveraging these models provides a scalable, computationally viable baseline for our exploratory analysis. 

This section is structured as follows. We first define ethos and pathos, and illustrate these rhetorical devices with manually annotated examples from OrigamIM (§\ref{ethos-pathos-definitions}). We then formalize the classification task (§\ref{task-formulation}) and outline the experimental setup (§\ref{experimental-setup}). We report classification results (§\ref{classification-results-origamim}) and assess label reliability through automatic and manual evaluation (§\ref{evaluation-reliability}).    

\subsubsection{Ethos and Pathos Definition} \label{ethos-pathos-definitions} \textit{Ethos} is the speaker's credibility and moral character, which are pivotal for persuasive communication \citep{Aristotle1991}. It is constructed through the speaker's demonstration of practical wisdom (\textit{phronesis}), virtue or moral excellence (\textit{arete}), and goodwill towards the audience (\textit{eunoia}). Ethos is not inherent but is established and perceived within the dynamic interaction between speaker and audience as the audience evaluates the speaker's trustworthiness and authority. Consequently, ethos functions as a rhetorical appeal that fundamentally relies on the audience's perception of the speaker's ethical character and expertise \citep{gajewska2024ethos,gajewska2024digital}. The goal of ethos annotation is to identify sentences where a speaker references the credibility or character of an entity, either in a supportive or attacking way.

\ex. \label{ex:ethosneg} \textcolor{ethosblue}{[Ethos: -]} \textit{Reading / watching \underline{CNN} is like eating junk food out of the trash bin.} 

\ex. \label{ex:ethospos} \textcolor{ethosblue}{[Ethos: +]} \textit{Man, \underline{she} is actually a really good politician if she can handle all that.} 

\textit{Pathos} refers to the second mode of persuasion that appeals to affective states of the audience \citep{Aristotle1991}. By strategically evoking specific emotions in the audience, the speaker influences their judgment and decision-making. The goal of pathos annotation is to capture rhetorical appeals that aim to induce specific affective states in the hearer rather than simply express emotions. 

\ex. \label{ex:pathosneg} \textcolor{pathosorange}{[Pathos: -]} \textit{I have a feeling there is going to be a HUGE genocide in the next 5-10 years.} 

\ex. \label{ex:pathospos} \textcolor{pathosorange}{[Pathos: +]} \textit{Look forward to see all the great things you will lead!} 

Together, these rhetorical modes illustrate persuasion as contingent on shifting interpersonal and emotional perspectives rather than objective, static truths.


\subsubsection{Task Formalization}\label{task-formulation} We define ethos and pathos classifiers as functions mapping textual inputs to categorical label sets: 
\begin{align*}
    \phi_{\text{ethos}}&: \mathcal{S} \to \mathcal{C_{\text{ethos}}}, & \phi_{\text{pathos}}: \mathcal{S} \to \mathcal{C_{\text{pathos}}}
\end{align*}
with $\mathcal{S}$ the space of all possible textual inputs, $\mathcal{C_{\text{ethos}}} = \{\textit{support}, \textit{attack}, \textit{neutral}\}$, and $\mathcal{C_{\text{pathos}}} = \{\textit{positive}, \textit{negative}, \textit{neutral}\}$. Since these label spaces are closely aligned, we unify them into a common scheme $\{+,-,0\}$, respectively. 
Ternary labeling reflects a principled trade-off between label granularity and reliability. While binary labeling would not cover the differentiation in non-neutral cases, increasing the number of labels is well known to reduce reliability. 

For each sentence $s$ and each interpretation $i$ in the interpretation-attitude pairs associated with $s$, denoted as $I_s = \{(i_1,a_1),...,(i_N,a_N)\}$, with $N$ the number of pairs, we assign: 
\begin{align*}
    et_s &= \phi_{\text{ethos}}(s), & pa_s &= \phi_{\text{pathos}}(s), \\
    et_{i_n} &= \phi_{\text{ethos}}(i_n), & pa_{i_n} &= \phi_{\text{pathos}}(i_n)
\end{align*}
as the ethos and pathos labels for the sentence $s$ and each interpretation $i_n$, respectively. 

\subsubsection{Experimental Setup}\label{experimental-setup} {We test with models finetuned on the respective sentence classification tasks and a pretrained large language model (LLM) for parameterizing functions $\phi_{\text{ethos}}$ and $\phi_{\text{pathos}}$.} 

\paragraph{Finetuned classifiers} {Following \citet{gajewska2024digital}, we establish baseline finetuned models for classifying ethos and pathos. To that end, we use the PolarIs dataset \citep{gajewska2024ethos}. It contains 15,588 sentences sourced from Reddit and X on topics such as COVID-19 vaccines, climate change, and U.S. presidential elections. Each sentence had been manually annotated by experts in rhetoric with ternary labels for ethos and pathos appeals. The label distribution for ethos is \{$+:854$; $-:2,610$; $0:12,124$\} and for pathos \{$+:895$; $-:4,329$; $0:10,365$\}. Inter-annotator agreement, measured with Cohen's $\kappa$ between an expert and trained annotators, indicates substantial agreement for ethos ($\kappa=0.78$) and moderate agreement for pathos ($\kappa=0.53$).}

{Fine-tuning is performed on 60\% of the dataset (training set). We conduct a random search across model architectures (BERT and RoBERTa) and key hyperparameters, including learning rate (1e-5--3e-4), batch size ([8, 16, 32, 64]), and number of training epochs (2-6). The best model for each classification task is selected based on F1 scores on the validation set (20\% of PolarIs data), i.e., RoBERTa-large (ethos) and RoBERTa-base (pathos). Hyperparameters configurations for reproducing the classification results are included in Appendix \ref{app:reproducibility}.}

\paragraph{LLM prompting} {Gemini 2.5 \citep{comanici2025gemini} independently performs ethos and pathos classification through in-context learning. For each classification task, we formulate three prompts that establish a zero-shot, few-shot, or chain-of-though (CoT) prompting setting. 
The prompting templates can be found in Appendix \ref{app:prompting_templates}.} 

\subsubsection{Classification Results for OrigamIM}\label{classification-results-origamim} 
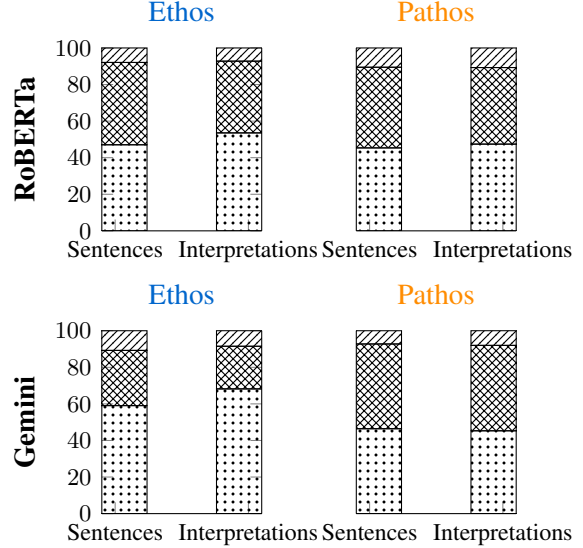
\begin{figure}[t!]
\centering
\begin{minipage}[t]{0.48\linewidth}
\centering
\begin{tikzpicture}
  \begin{axis}[
    ybar stacked,
    bar width=24pt,
    width=\linewidth,
    height=4cm,
    ymin=0, ymax=100,
    symbolic x coords={Sentences, Interpretations},
    xtick=data,
    xticklabel style={font=\footnotesize},
    yticklabel style={font=\footnotesize},
    axis x line*=bottom,
    ylabel={\textbf{RoBERTa}},
    title={\textcolor{ethosblue}{Ethos}},
  ]
    \addplot+[draw=black, fill=none, pattern=dots] coordinates {(Sentences,47.06) (Interpretations,53.55)};
    \addplot+[draw=black, fill=none, pattern=crosshatch] coordinates {(Sentences,45.01) (Interpretations,39.23)};
    \addplot+[draw=black, fill=none, pattern=north east lines] coordinates {(Sentences,7.93) (Interpretations,7.22)};
  \end{axis}
\end{tikzpicture}
\end{minipage}
\hfill
\begin{minipage}[t]{0.48\linewidth}
\centering
\begin{tikzpicture}
  \begin{axis}[
    ybar stacked,
    bar width=24pt,
    width=\linewidth,
    height=4cm,
    ymin=0, ymax=100,
    symbolic x coords={Sentences, Interpretations},
    xtick=data,
    xticklabel style={font=\footnotesize},
    title={\textcolor{pathosorange}{Pathos}},
    ytick=\empty,                
    axis x line*=bottom,         
    ymajorgrids=false,
  ]
    \addplot+[draw=black, fill=none, pattern=dots] coordinates {(Sentences,45.35) (Interpretations,47.43)};
    \addplot+[draw=black, fill=none, pattern=crosshatch] coordinates {(Sentences,44.10) (Interpretations,41.86)};
    \addplot+[draw=black, fill=none, pattern=north east lines] coordinates {(Sentences,10.55) (Interpretations,10.71)};

    \node[anchor=west, font=\small, inner sep=1pt, xshift=4pt]
        at (axis cs:Interpretations,23.73) {neutral};
    \node[anchor=west, font=\small, inner sep=1pt, xshift=4pt]
        at (axis cs:Interpretations,68.395) {negative};
    \node[anchor=west, font=\small, inner sep=1pt, xshift=4pt]
        at (axis cs:Interpretations,94.685) {positive};
        
  \end{axis}
\end{tikzpicture}
\end{minipage}
\hfill
\begin{minipage}[t]{0.48\linewidth}
\centering
\begin{tikzpicture}
  \begin{axis}[
    ybar stacked,
    bar width=24pt,
    width=\linewidth,
    height=4cm,
    ymin=0, ymax=100,
    symbolic x coords={Sentences, Interpretations},
    xtick=data,
    xticklabel style={font=\footnotesize},
    yticklabel style={font=\footnotesize},
    axis x line*=bottom,
    ylabel={\textbf{Gemini}},
    title={\textcolor{ethosblue}{Ethos}},
  ]
    \addplot+[draw=black, fill=none, pattern=dots] coordinates {(Sentences,59.05) (Interpretations,68.19)};
    \addplot+[draw=black, fill=none, pattern=crosshatch] coordinates {(Sentences,30.19) (Interpretations,23.32)};
    \addplot+[draw=black, fill=none, pattern=north east lines] coordinates {(Sentences,10.76) (Interpretations,8.50)};
  \end{axis}
\end{tikzpicture}
\end{minipage}
\hfill
\begin{minipage}[t]{0.48\linewidth}
\centering
\begin{tikzpicture}
  \begin{axis}[
    ybar stacked,
    bar width=24pt,
    width=\linewidth,
    height=4cm,
    ymin=0, ymax=100,
    symbolic x coords={Sentences, Interpretations},
    xtick=data,
    xticklabel style={font=\footnotesize},
    title={\textcolor{pathosorange}{Pathos}},
    ytick=\empty,                
    axis x line*=bottom,         
    ymajorgrids=false,
  ]
    \addplot+[draw=black, fill=none, pattern=dots] coordinates {(Sentences,46.41) (Interpretations,45.29)};
    \addplot+[draw=black, fill=none, pattern=crosshatch] coordinates {(Sentences,46.35) (Interpretations,46.70)};
    \addplot+[draw=black, fill=none, pattern=north east lines] coordinates {(Sentences,7.24) (Interpretations,8.02)};

    \node[anchor=west, font=\small, inner sep=1pt, xshift=4pt]
        at (axis cs:Interpretations,23.73) {neutral};
    \node[anchor=west, font=\small, inner sep=1pt, xshift=4pt]
        at (axis cs:Interpretations,68.395) {negative};
    \node[anchor=west, font=\small, inner sep=1pt, xshift=4pt]
        at (axis cs:Interpretations,94.685) {positive};
        
  \end{axis}
\end{tikzpicture}
\end{minipage}

\caption{Ethos (left) and Pathos (right) label distributions in OrigamIM (percentages), with neutral ($0$, dots), negative ($-$, crosshatches), and positive labels ($+$, lines). The panel on top show the predicted distributions for RoBERTa, the panel at the bottom those for Gemini.}
\label{fig:label_distributions_origamim}
\end{figure}

{Neutral labels dominate for both ethos and pathos, followed by negative and positive. This trend is consistent across both classification methods (see Figure \ref{fig:label_distributions_origamim}).} 

\subsubsection{Evaluation of Label Reliability}\label{evaluation-reliability}

{We assess the reliability of the silver-standard ethos and pathos labels for the OrigamIM dataset \textit{automatically} and \textit{manually}.} 
{\paragraph{Automatic evaluation on gold standard external data} We evaluate classification performance on two existing ethos/pathos datasets: PolarIs (held-out test set; three labels; 20\% of the dataset) \citep{gajewska2024ethos} and EU DisinfoTest \citep{sosnowski2024eu} (two labels). The latter dataset contains 1,344 narratives, which span a single or multiple sentences, on a variety of disinformation-sensitive topics such as migration and health. It provides ethos and pathos-heavy reformulations of the narratives produced by GPT-4. Ethos and pathos labeling is therefore binary, i.e., ethos/non-ethos and pathos/non-pathos, with 1,344 samples in each class. Likewise, we assess performance on PolarIs in a binary labeling setting, for which we aggregate the positive and negative labels into a joint class. This allows us to analyze the domain transferability of the classifiers. It is important to note that annotation in the PolarIs dataset regards others-directed ethos, while in EU DisinfoTest it regards self-referential ethos, which might lead to systematic performance differences due to the distinct pragmatic realizations of credibility cues across these datasets.} 

Table \ref{tab:automatic-validation} reports the precision, recall, and F1-scores (mean and standard deviation) over three runs. Performance of Gemini closely aligns with high inter-annotator agreement for ethos ($\kappa=0.78$) and a moderate agreement for pathos ($\kappa=0.53$) reported for PolarIs, reflecting the greater interpretive ambiguity of emotional appeals. Fine-tuned RoBERTa shows a comparable performance for both classification tasks (F1=0.88-0.91), suggesting that in-domain training allows the model to approximate expert-level consistency on PolarIs. The performance drop in a domain-transfer condition indicates that these results partly reflect dataset-specific regularities. However, this result is expected since different annotation guidelines have been used for the two datasets, so the models are likely challenged by differences in perspective (self- vs. other-directed ethos), rhetorical framing, and topic distribution. 


\begin{table}[!t]
    \centering
    \label{tab:ethos-pathos}
    \footnotesize

    \begin{subtable}[t]{\linewidth}
        \centering
        \label{tab:ethos}
        \begin{tabular}{p{2cm}ccc}
             \toprule
            \rowcolor{blue!20}\textbf{Model} & \textbf{Precision} & \textbf{Recall} & \textbf{F1} \\
            \midrule
            \rowcolor{gray!20}\multicolumn{4}{c}{\textit{PolarIs - 3 labels}} \\
            RoBERTa    & $0.91_{\pm 0.01}$ & $0.91_{\pm 0.01}$ & $0.91_{\pm 0.01}$\\
            Gemini 2.5 &      $0.79_{\pm 0.01}$  &   $0.77_{\pm 0.01}$  &  $0.78_{\pm 0.01}$        \\
            \rowcolor{gray!20} \multicolumn{4}{c}{\textit{PolarIs - 2 labels}} \\
            RoBERTa    &     $0.93_{\pm 0.01}$               &  $0.93_{\pm 0.01}$                &      $0.93_{\pm 0.01}$         \\
            Gemini 2.5 &  $0.80_{\pm 0.01}$  &   $0.80_{\pm 0.01}$  &  $0.80_{\pm 0.01}$  \\
            \rowcolor{gray!20} \multicolumn{4}{c}{\textit{EU DisinfoTest - 2 labels}} \\
            RoBERTa    &    $0.79_{\pm 0.01}$            &           $0.56_{\pm 0.01}$     &               $0.64_{\pm 0.01}$   \\
            Gemini 2.5 &       $0.87_{\pm 0.05}$         &          $0.49_{\pm 0.05}$     &            $0.57_{\pm 0.05}$     \\
            \bottomrule
        \end{tabular}
        \caption{Ethos classification.}
    \end{subtable}
    \hfill
    \begin{subtable}[t]{\linewidth}
        \centering
        \label{tab:pathos}
        \begin{tabular}{p{2cm}ccc}
             \toprule
            \rowcolor{orange!20}\textbf{Model} & \textbf{Precision} & \textbf{Recall} & \textbf{F1} \\
             \midrule
            \rowcolor{gray!20}\multicolumn{4}{c}{\textit{PolarIs - 3 labels}} \\
            RoBERTa    & $0.88_{\pm 0.01}$ & $0.88_{\pm 0.01}$ & $0.88_{\pm 0.01}$\\
            Gemini 2.5 &          $0.67_{\pm 0.00}$        &            $0.64_{\pm 0.01}$     &     $0.64_{\pm 0.01}$         \\
            \rowcolor{gray!20}\multicolumn{4}{c}{\textit{PolarIs - 2 labels}} \\
            RoBERTa    &      $0.90_{\pm 0.01}$         &     $0.90_{\pm 0.01}$            &       $0.90_{\pm 0.01}$     \\
            Gemini 2.5 &    $0.69_{\pm 0.00}$       &            $0.68_{\pm 0.01}$     &     $0.67_{\pm 0.01}$           \\
            \rowcolor{gray!20}\multicolumn{4}{c}{\textit{EU DisinfoTest - 2 labels}} \\
            RoBERTa    &      $0.85_{\pm 0.01}$           &           $0.50_{\pm 0.01}$     &               $0.54_{\pm 0.01}$     \\
            Gemini 2.5 &      $0.86_{\pm 0.00}$          &           $0.47_{\pm 0.01}$    &           $0.51_{\pm 0.01}$     \\
            \bottomrule
        \end{tabular}
        \caption{Pathos classification.}
    \end{subtable}
    \caption{\textbf{Automatic} validation of ethos/pathos label reliability: Performance on ethos (top) and pathos (bottom) classification across datasets.}
    \label{tab:automatic-validation}
\end{table}

Analysis of inconsistent classification by the two approaches in EU DisInfo indicates that differences in perspective are primarily responsible for this disparate performance. Let's consider the following examples: 
\ex. \label{ex:stiglitz} Nobel Laureate Joseph Stiglitz has expressed concerns that the euro restricts economic policy options for member states, suggesting that countries lose economic autonomy upon adopting the euro. Moreover, the International Monetary Fund has documented persistent economic struggles in countries post-euro adoption, indicating a trend towards economic decline. 

\ex. \label{ex:failure} The euro and Eurozone policies have been a failure, leading to increased economic disparities and hardships for member countries. 

The divergence in classification stems from a fundamental mismatch in how the two models operationalize the ethos dimension based on their respective training sets. The model trained on the PolarIs dataset is optimized to recognize other-directed ethos appeals, such as the explicit reputation attacks on institutional policies seen in example \ref{ex:failure}\footnote{``Individuals, groups or organizations that participate in a discussion or are just a third party that is mentioned in the exchanges could be targets of
such appeals.'' \citep[p. 2]{gajewska2024ethos}}. Conversely, the model exposed to the EU DisInfo guidelines is calibrated to capture instances where authority is outsourced to external entities, specifically flagging arguments from expert opinion like the citations of a Nobel Laureate in example \ref{ex:stiglitz}\footnote{``Ethos is used to convey the writer’s credibility and authority.'' \citep[p. 14718]{sosnowski2024eu} }. Consequently, when deployed across domains, the models fail to generalize because they are searching for fundamentally different rhetorical mechanisms: one tracks direct interpersonal or institutional confrontation, while the other tracks the structural integration of external credibility. This result confirms that the domain-transfer degradation is a structural misalignment in perspective modeling rather than mere lexical noise.

\paragraph{Manual evaluation on gold standard OrigamIM data} We further validate label reliability \textit{manually} on a randomly selected subset of 100 sentences and interpretations from OrigamIM. {Two in-house expert annotators independently establish gold standard ethos and pathos labels for these samples, for which they follow the annotation guidelines used for constructing PolarIs (annotation details in Appendix \ref{app:annotation_guidelines_validation}). Agreement between the two raters has been achieved in 71\% of cases for ethos ($\kappa=0.5$) and in 69\% of cases for pathos ($\kappa=0.4$). Note that these $\kappa$ values likely underestimate true inter-annotator reliability, as several label categories are sparsely represented in this small sample (some occurring only 5-10 times), which is known to lower Cohen’s $\kappa$. Due to the high cost and time requirements of expert annotation, we opted for a focused manual evaluation between the two individuals. 
To obtain the final labels, disagreements were resolved using raters' confidence levels for individual cases.} 

\begin{table}[!t]
    \centering
    \label{tab:ethos-pathos2}
    \footnotesize

    \begin{subtable}[t]{\linewidth}
        \centering
        \label{tab:ethos2}
        \begin{tabular}{p{2cm}ccc}
             \toprule
        \rowcolor{blue!20}\textbf{Model} & \textbf{Precision} & \textbf{Recall} & \textbf{F1} \\
            \midrule
            \rowcolor{gray!20} \multicolumn{4}{c}{\textit{OrigamIM - 3 labels}} \\
            RoBERTa    &     0.75             &  0.63                 &      0.64         \\
            Gemini 2.5 &  0.76           &     0.71       &                 0.73 \\
            \bottomrule
        \end{tabular}
        \caption{Ethos classification.}
    \end{subtable}
    \hfill
    \begin{subtable}[t]{\linewidth}
        \centering
        \label{tab:pathos2}
        \begin{tabular}{p{2cm}ccc}
             \toprule
            \rowcolor{orange!20}\textbf{Model} & \textbf{Precision} & \textbf{Recall} & \textbf{F1} \\
             \midrule
            \rowcolor{gray!20}\multicolumn{4}{c}{\textit{OrigamIM - 3 labels}} \\
            RoBERTa    &      0.73        &     0.66            &       0.67   \\
            Gemini 2.5 &    0.75        &            0.68     &     0.70           \\
            \bottomrule
        \end{tabular}
        \caption{Pathos classification.}
    \end{subtable}
    \caption{\textbf{Manual} validation of ethos/pathos label reliability: Performance on ethos (top) and pathos (bottom) classification on a subset of the OrigamIM dataset.}
    \label{tab:manual-validation}
\end{table}

Table \ref{tab:manual-validation} reports model performance on the gold-standard subset of OrigamIM. For ethos, Gemini 2.5 outperforms RoBERTa (F1 = 0.73 vs. 0.64), mainly due to higher recall, indicating better coverage of credibility-related cues in this setting. For pathos, both models achieve comparable performance, with a modest advantage of Gemini 2.5 (F1 = 0.70 vs. 0.67), suggesting a slightly more robust alignment with human judgments on emotional appeals. Eventually, we combine RoBERTa and Gemini decisions through a majority voting scheme for all downstream data analyses.


\section{Analysis}
This section analyzes the OrigamIM data to examine how ethos and pathos expressed in an original sentence are reflected in audience interpretations. 

\begin{figure*}[ht!]
\centering
\begin{minipage}[t]{0.48\textwidth}
\footnotesize
\raggedright
\begin{tcolorbox}[colback=lightgray,colframe=black!20,arc=3pt,boxrule=0.5pt]
\textbf{[\textit{Complete Divergence}]}\\[4pt]
\textbf{Sentence:} \\
\textcolor{ethosblue}{[Ethos: +]} \textcolor{pathosorange}{[Pathos: +]}\\
\textit{It's a celebration of the white spirit and the lack of racism in so many white people.}\\[6pt]
\textbf{Interpretations:} \\
\textcolor{ethosblue}{[Ethos: -]} \textcolor{pathosorange}{[Pathos: -]} The author is making the frankly bizarre claim that Juneteenth is a holiday about how some white people aren't racist. \\
\textcolor{ethosblue}{[Ethos: 0]} \textcolor{pathosorange}{[Pathos: 0]} People shouldn't be angry at Juneteenth. \\
\textcolor{ethosblue}{[Ethos: 0]} \textcolor{pathosorange}{[Pathos: -]} Many whites celebrate the enthusiasm of white spirit with no regards to racism. \\
\textcolor{ethosblue}{[Ethos: 0]} \textcolor{pathosorange}{[Pathos: 0]} Juneteenth celebrates the role white people played in the emancipation of American slaves. \\
\textcolor{ethosblue}{[Ethos: 0]} \textcolor{pathosorange}{[Pathos: -]} It's a celebration of the lack of racism white people. \\
\end{tcolorbox}
\end{minipage}\hfill
\begin{minipage}[t]{0.48\textwidth}
\footnotesize
\raggedright
\begin{tcolorbox}[colback=lightgray,colframe=black!20,arc=3pt,boxrule=0.5pt]
\textbf{[\textit{Full Alignment}]}\\[4pt]
\textbf{Sentence:} \\
\textcolor{ethosblue}{[Ethos: 0]} \textcolor{pathosorange}{[Pathos: 0]}\\
\textit{They are essentially starting at \$0.}\\[6pt] \\
\textbf{Interpretations:} \\
\textcolor{ethosblue}{[Ethos: 0]} \textcolor{pathosorange}{[Pathos: 0]} The writer states that people who do not have student debt or a degree are starting at a worse point than people with student debt and a degree. \\
\textcolor{ethosblue}{[Ethos: 0]} \textcolor{pathosorange}{[Pathos: 0]} Compensation for payments towards debt would be neutral. \\
\textcolor{ethosblue}{[Ethos: 0]} \textcolor{pathosorange}{[Pathos: 0]} People who clear their student loans have no money left with them at that point. \\
\textcolor{ethosblue}{[Ethos: 0]} \textcolor{pathosorange}{[Pathos: 0]} Their loan has been cancelled. \\
\textcolor{ethosblue}{[Ethos: 0]} \textcolor{pathosorange}{[Pathos: 0]} They are starting at \$0. \\ \\ \\

\end{tcolorbox}
\end{minipage}

\caption{Examples of complete divergence (\textit{left}) and full alignment (\textit{right}) in ethos and pathos labels between the original sentence (\textit{top}, in italics) and corresponding audience interpretations (\textit{bottom}).}
    \label{tab:cases2}
\end{figure*}

\begin{tcolorbox}[colback=lightgray,colframe=black!50,boxrule=0.5pt,arc=4pt]
\textbf{RQ1: Are ethos and pathos in the original sentence reflected differently in audience interpretations?}
\end{tcolorbox}

When comparing the labels between sentence and interpretation, we observe that the interpretations do not consistently align with the original sentence in terms of ethos (i.e., $et_s \neq et_{i_n}$) and pathos (i.e., $pa_s \neq pa_{i_n}$). {Only $74.2\%$ and $70.4\%$ of interpretations match the original sentence's ethos and pathos labels, respectively. Full alignment, where all interpretations of a given sentence reflect both the same ethos and pathos labels as the original, occurs in just 15.9\% of the sentences. By contrast, complete divergence, where all interpretations differ from the original sentence in both dimensions, is rare, appearing in only one case.} Figure \ref{tab:cases2} illustrates examples of full alignment and complete divergence. 
These findings show that rhetorical signals in a sentence are often perceived differently, highlighting the interpretive variability of ethos and pathos among the audience.

\begin{figure}[t!]
\centering
\begin{subfigure}[t]{0.48\textwidth}
\centering
\begin{tikzpicture}
\begin{axis}[
    ybar,
    bar width=9pt,
    enlarge x limits=0.3,
    ylabel={Matching rate (\%)},
    symbolic x coords={all, neutral, non-neutral, baseline-p, neutral-pa, not-neutral-pa},
    ylabel style={font=\scriptsize},
    y tick label style={font=\scriptsize},
    xtick=data,
    nodes near coords,
    ymin=0,
    ymax=100,
    x tick label style={rotate=45, anchor=east,font=\scriptsize},
    legend style={at={(0.56,-0.2)},anchor=west,legend columns=-1,font=\scriptsize},
    width=8cm,
    height=5cm,
    every node near coord/.append style={font=\scriptsize}
]
\addplot+[fill=ethosblue, nodes near coords={
    \ifnum\coordindex=0 {} 
    \else\ifnum\coordindex=1 {{\textbf{+11.3\%}}} 
    \else\ifnum\coordindex=2 {\textcolor{red}{\textbf{-22.3\%}}} 
    \fi\fi\fi
}] coordinates {
    (all,74.3) 
    (neutral,85.6) 
    (non-neutral,51.0)
};

\addplot+[fill=pathosorange, nodes near coords={
    \ifnum\coordindex=0 {} 
    \else\ifnum\coordindex=1 {{\textbf{+2.2\%}}} 
    \else\ifnum\coordindex=2 {\textcolor{red}{\textbf{-7.7\%}}} 
    \fi\fi\fi
}] coordinates {
    (baseline-p,70.4) 
    (neutral-pa,72.6) 
    (not-neutral-pa,62.7)
};

\legend{Ethos, Pathos}
\end{axis}
\end{tikzpicture}
\caption{Matching rates between sentence and interpretation based on the sentence label for ethos (left) and pathos (right).}
\label{fig:rq1matchoverall}
\end{subfigure}
\hfill
\begin{subtable}[t]{0.48\textwidth}
\centering
\footnotesize
\begin{tabular}{p{0.7cm}p{1.1cm}p{2.05cm}p{2.05cm}}
\toprule
\textbf{Label for $s$} & \textbf{\# Labels in $I_s$} & \textbf{\% \textcolor{ethosblue}{[Ethos]}} & \textbf{\% \textcolor{pathosorange}{[Pathos]}} \\
\midrule
\textbf{$0$} & 1 / 2 / 3 & \textbf{67.2} / 31.0 / 1.7 & \textbf{50.4} / 45.7 / 3.9 \\
\textbf{$-$} & 1 / 2 / 3 & 30.3 / \textbf{59.2} / 10.5 & 27.0 / \textbf{62.0} / 10.9 \\
\textbf{$+$} & 1 / 2 / 3 & 35.5 / \textbf{62.3} / 2.2 & 37.4 / \textbf{60.1} / 2.5 \\
\bottomrule
\end{tabular}
\caption{Share of interpretation sets with 1, 2, or 3 distinct ethos/pathos labels, conditioned on the original sentence label}
\label{tab:combined_ethos_pathos_sidebyside}
\end{subtable}

\caption{Overview: (a) Matching rates by neutrality; (b) Distribution of number of labels in interpretations.}
\label{fig:plot_table_side_by_side}
\end{figure}

\begin{tcolorbox}[colback=lightgray,colframe=black!50,boxrule=0.5pt,arc=4pt]
\textbf{RQ2: Which ethos and pathos labels trigger greater variability in audience interpretations?}
\end{tcolorbox}

Original sentences with neutral ethos ($et_s = 0$) and pathos ($pa_s = 0$) are associated with significantly higher alignment in audience interpretations compared to non-neutral cases, with statistical significance at $p<0.001$ for ethos and $p<0.001$ for pathos (tested using the $\chi^2$ statistic); see Figure \ref{fig:rq1matchoverall}. This suggests that rhetorical neutrality facilitates consensus among readers. 
Sentences expressing ethos support ($et_s = +$) and positive pathos ($pa_s = +$) show the highest variability (see Figure \ref{tab:combined_ethos_pathos_sidebyside}). We further observe that the same ethos and pathos label is expressed in all interpretations for only 36\% and 28\% of sentences, respectively. 
These findings indicate that rhetorically charged content tends to elicit a broader range of audience interpretations, underscoring the interpretive flexibility and complexity inherent in emotionally loaded discourse.

\begin{table}[h!]
\centering
\setlength{\tabcolsep}{3pt} 
\begin{tabular}{lrrrr}
\hline
\textbf{Predictor} & \textbf{$\beta$} & \textbf{SE} & \textbf{$t$} & \textbf{$p$} \\
\hline
Intercept & 1.54 & 0.02 &  87.37 & *** \\
\textcolor{pathosorange}{[Pathos: +]} & 0.30 & 0.05 & 6.02 & ***  \\
\textcolor{pathosorange}{[Pathos: -]} & 0.11 & 0.03 & 4.50 & *** \\ 
\hline

Intercept & 1.35 & 0.02 & 92.55 & *** \\
\textcolor{ethosblue}{[Ethos: +]} & 0.46 & 0.05 & 10.26 &  *** \\
\textcolor{ethosblue}{[Ethos: -]} & 0.32 & 0.03 & 12.41 & *** \\
\hline
\end{tabular} 
\caption{OLS regression predicting \textbf{variability in interpretation labels from sentence labels}. Neutral labels are set as baselines. Statistical significance: *** ($p<.001$).}
\label{tab:regression_variability}
\end{table}

Table \ref{tab:regression_variability} presents the results of two OLS regression models predicting variability in interpretation labels from sentence-level rhetorical appeals. The intercepts represent the expected level of interpretive variability for sentences without marked pathos or ethos appeals. In the first model, pathos appeals are significantly associated with increased variability in sentence interpretations. Sentences containing positive pathos exhibit higher interpretive variability compared to neutral sentences ($\beta = 0.30$, $p < .001$), indicating that emotionally positive content elicits more divergent audience interpretations. 

Notably, negative pathos is also associated with a significant increase in variability ($\beta = 0.11$, $p < .001$), although the magnitude of this effect is smaller. The second model focuses on ethos appeals and reveals even stronger effects. Both positive and negative ethos are associated with statistically significant increases in interpretive variability relative to neutral sentences: positive ($\beta = 0.46$, $p < .001$) and negative ($\beta = 0.32$, $p < .001$). This pattern indicates that credibility-related cues, whether reinforcing or undermining someone’s authority, markedly amplify disagreement or divergence in sentence interpretations.

\begin{tcolorbox}[colback=lightgray,colframe=black!50,boxrule=0.5pt,arc=4pt]
\textbf{RQ3: Can ethos and pathos in the original sentence predict attitudes toward speakers?}
\end{tcolorbox}

We examine whether ethos and pathos in the sentence, i.e., $et_s$ and $pa_s$, can predict a reader's attitude towards the author, $a_i$. To that end, linear regression tested the effects of ethos and pathos on audience attitude ratings, treating attitudes as continuous numerical values. The model is significant for both pathos ($F(2, 9848) = 72.34, p < .001$) and ethos ($F(2, 9846) = 25.29, p < .001$). Table \ref{tab:regression} indicate that sentences with positive pathos/ethos elicit more positive attitudes than those with neutral pathos/ethos. On the other hand, sentences with negative pathos/ethos trigger more negative attitudes than those with neutral pathos/ethos. 
The linear regression models for predicting \textit{variability} in reader attitudes within the interpretation set accompanying a sentence, $I_s$, using sentence-level ethos and pathos are not statistically significant. 
Note that the limited size of $I_s$ potentially biases attitude variability {and the statistical findings regarding attitude variability.} 

\begin{table}[t!]
\centering
\setlength{\tabcolsep}{3pt} 
\begin{tabular}{lrrrr}
\hline
\textbf{Predictor} & \textbf{$\beta$} & \textbf{SE} & \textbf{$t$} & \textbf{$p$} \\
\hline
Intercept & 3.02 & 0.01 & 208.55 & *** \\
\textcolor{pathosorange}{[Pathos: +]} & 0.37 & 0.04 & 8.89 & ***  \\
\textcolor{pathosorange}{[Pathos: -]} & -0.12 & 0.02 & -5.78 & *** \\ \hline
Intercept & 3.00 & 0.01 & 234.26 & *** \\ 
\textcolor{ethosblue}{[Ethos: +]} &  0.18 & 0.04 & 4.62 &  *** \\
\textcolor{ethosblue}{[Ethos: -]} & -0.10 & 0.02 & -4.46 & *** \\
\hline
\end{tabular} 
\caption{{OLS regression predicting \textbf{audience attitudes from sentence-level pathos and ethos appeals}. Neutral labels are set as baselines. Statistical significance: *** ($p<.001$).}}
\label{tab:regression}
\end{table}

\begin{table*}[h]
    \centering
    \footnotesize
    \begin{tabular}{p{8.85cm}p{3.1cm}p{1cm}p{1.25cm}} \hline
      \textbf{Example sentence} &   \textbf{Category} & \textbf{\# Cases} & \textbf{\% Cases} \\ \hline
\footnotesize{\textit{Sure, I do disagree with her on a lot but I think she's reasonable enough to work with people and not accuse them of being a Russian asset.}} &  \textbf{Author opinion}         &  64  & 28.6 \\ \addlinespace[0.1cm]
\footnotesize{\textit{I've heard Christians talk about how it's actually good when someone's uncomfortable: it means that they're getting closer to actually having a breakthrough of faith or realizing the error in their non-christian ways.}} &  \textbf{Value system}            & 44  & 19.6 \\ \addlinespace[0.1cm]
\footnotesize{\textit{Hell, Trump's done a lot of damage to the USA's reputation just because he tweets garbage non-stop.}} &  \textbf{Political partisanship}  & 36  & 16.1 \\
\footnotesize{\textit{Black Lives Matter (as practiced by white people) DOES imply that white lives DON'T matter as much.}}  &  \textbf{Social identity/position}         &  30 & 13.4  \\
\footnotesize{\textit{She gets paid every month for something, and it's never going to end.}}  &  \textbf{Out of context}          & 20  & 8.9  \\  \addlinespace[0.1cm]
\footnotesize{\textit{Where are the federal agents going door to door telling people to not smoke tobacco and avoid high sugar foods?}}  &  \textbf{Rhetorical question}     & 16  & 7.1  \\
\footnotesize{\textit{I had discussions with some black people and (for lack of a better term) "woke" white people I know.}}  &  \textbf{Personal experience} & 10 & 4.5  \\
\footnotesize{\textit{but I promise this isn't a 'white people are the real victims' rant and there will be new questions in here.}}  &  \textbf{Sarcasm}                  & 4 & 1.8  \\
         \hline
    \end{tabular}
    \caption{Factors influencing variability of ethos and pathos labels between sentences and interpretations. }
    \label{tab:manualanalysis}
\end{table*}

\begin{tcolorbox}[colback=lightgray,colframe=black!50,boxrule=0.5pt,arc=4pt]
{\textbf{RQ4: Which factors beyond ethos and pathos in the sentence influence interpretation variability?}}
\end{tcolorbox}
{
We manually analyze 100 randomly sampled cases, where at least one label differs between the interpretation and the sentence plus all cases, where all three ethos and pathos labels (\{+, -, 0\}) are present in a set of interpretations: there are 51 such sentences for ethos and 76 for pathos.}
{Our analysis reveals eight factors that may explain why audience interpretations of the same sentence differ, alongside ethos and pathos (Table \ref{tab:manualanalysis}). These factors could be seen as interpreter-related, content- and form-based, and contextual factors. 
\paragraph{Interpreter-related factors:} These arise from the background, beliefs, and perspectives of the interpreter rather than from the sentence itself. This group includes \textit{value system differences}, \textit{political partisanship}, \textit{social identity/position}, and \textit{personal experience}, all of which reflect how readers project their own moral frameworks, identities, and lived experiences onto the content.
\paragraph{Content- and form-based factors:} These are driven by linguistic or rhetorical properties of the sentence that invite multiple readings. This category covers \textit{author opinion} (e.g., first- vs. third-person framing), \textit{rhetorical questions}, and \textit{sarcasm}, where stylistic or pragmatic cues alter perceived commitment, intent, or emotional tone. 
\paragraph{Contextual factors:} They stem from missing or underspecified contextual information. Particularly, \textit{out-of-context} effects lead interpreters to infer referents, stance, or targets differently, resulting in divergent ethos and pathos assignments. 

Extended definitions of these factors can be found in Appendix \ref{app:definition_clarification}.}
\section{Discussion} 

Our results align with theoretical accounts that emphasize mechanisms sustaining interpretive divides in digital communication. The notion of \textit{echo chambers}, for instance, explains how selective exposure to ideologically homogeneous content reinforces prior beliefs and limits engagement with opposing views \citep{garimella2018political, del2016echo}. Within such environments, ethos cues may be readily accepted when they align with in-group norms but dismissed or inverted when attributed to out-group speakers. 

Furthermore, the observed relation between rhetorical charge and interpretation variance suggests a testable hypothesis: content with high rhetorical load may be more prone to divergent interpretations and conflict. This distinction raises the possibility of investigating ``interpretive instability'' as a separate dimension from traditional toxicity that triggers hostility. Such an approach could deepen our understanding of how persuasion interacts with audience cognition and inform research on the dynamics of conflict and consensus in online discourse.

These findings present critical implications for the computational modeling of discourse and pragmatic interpretation in multi-party interaction. By demonstrating a 30\% divergence between original text and reader interpretations, this work challenges the widespread assumption that textual content can serve as a direct proxy for audience reception. Instead, the heightened variability observed in rhetorically charged content indicates that ground truth in communication data is inherently unstable, particularly when dealing with persuasive appeals like ethos and pathos. For the NLP community, this means that optimizing language models purely on semantic similarity metrics ignores a massive chasm in actual human communication. To model social phenomena like polarization, we must pivot from modeling what the text says to modeling the distribution of what the text evokes. This challenges the field to develop a new class of receiver-centric pragmatic models.

\section{Conclusion}
The paper presented a computational study on the transformation of rhetorical appeals (specifically ethos and pathos) within the interpretive space of social media discourse. Rather than viewing rhetoric as a static property of the text, we quantified the extent to which these modes of persuasion are preserved or distorted during the interpretation process. Our methodology relied on a statistical examination of the alignment between source and target rhetorical labels, employing regression analyses to isolate the specific effects of credibility and emotional appeals on interpretive variability and the formation of audience attitudes. 

Our findings show that rhetorical appeals are significant predictors of interpretive divergence: while rhetorical neutrality fosters consensus, marked ethos and pathos appeals actively catalyze disagreement in meaning construction. Furthermore, we demonstrated that these appeals can predict reader attitudes toward the author, highlighting the direct link between rhetorical strategy and social perception. By disentangling the linguistic and cognitive factors that mediate this variation, such as value systems and political affiliation, this work advances a perspectivist approach to analyzing rhetorical influence. 

While this study establishes a baseline for modeling rhetorical interpretation within the silent audience, it also opens several avenues for future empirical investigation. First, while we isolate ethos and pathos as primary drivers of interpretative variance, future work expands the feature space to account for potentially correlated alternative dimensions. Specifically, we plan to disentangle rhetorical appeals from confounding variables such as ideological alignment, text complexity, and implicit topic bias, which may co-vary with ethos and pathos and contribute to the observed semantic divergence. Second, a vital next step is to bridge the gap between the latent cognitive states of the silent audience and the overt behaviors of active users. We intend to collect larger datasets that pair human written interpretations with platform-level engagement signals, such as upvotes, shares, and comment counts. Cross-referencing these behavioral metrics with our interpretation-level rhetorical patterns will allow us to determine whether the subtle shifts in audience perception we uncovered correlate with or diverge from high engagement digital discourse.

\section*{Limitations}
The design of the study constrains the scope and robustness of its findings. 

The analyses rely exclusively on the OrigamIM dataset, which contains English sentences collected from Reddit posts from the r/ChangeMyView subreddit and pairs each sentence with a set of five human-written interpretations. The single-language property of the sentences limits the generalizability of results to other languages. Moreover, the size of the interpretation set is too small to fully capture all possible audience interpretations of a given sentence. For this, a much larger set coming from a diverse group of readers is necessary to approximate true variation in sentence interpretations, which is, to our knowledge, non-existing.

{Nonetheless, we contend that our findings possess a high degree of generalizability despite the data originating from a specific subreddit. That is because the underlying architecture of most social media platforms is fundamentally the same: they serve as digital arenas for the expression and exchange of personal opinions. By using the core feature of source opinions and their subsequent interpretations from Reddit, we are analyzing a basic communicative unit, i.e., the message and its reception, that is a universal characteristic of social media interaction. While platform-specific norms vary, the cognitive mechanisms involved in processing rhetorically charged content, such as the projection of value systems or partisan filters, are structural features of human communication that transcend any single platform.}

Ethos and pathos classification depends on finetuned RoBERTa-based models, which, despite reasonable performance, are subject to noise and moderate error rates inherent in subjective labeling tasks. The downstream analyses are based on statistically significant patterns rather than individual predictions such that observed trends reflect meaningful rhetorical phenomena rather than random noise. Nonetheless, the predicted labels should be interpreted as model-predicted rhetorical signals rather than absolute truths. 



\section*{Ethical Considerations}
This research involves analysis of social media text, raising several ethical considerations. First, the dataset used (OrigamIM) is publicly available and licensed for research (CC BY-SA 4.0), used solely according to intended purposes with consent from the dataset creators, thereby respecting data privacy and licensing norms. The subjective nature of rhetorical interpretation, influenced by cultural background and prior beliefs, may lead to overgeneralization or exclusion of minority viewpoints if models are deployed without careful contextualization. Furthermore, findings concerning polarization effects could be misapplied for manipulative purposes such as targeted disinformation. To mitigate harm, the work emphasizes interpretive variability and perspectivism, cautioning against simplistic or reductionist deployment of classification results.

\section*{Acknowledgments}
This work has been funded in part by the Research Foundation - Flanders (FWO) under grant G0L0822N through the CHIST-ERA project ``iTRUST Interventions against Polarisation in Society for Trustworthy Social Media. From Diagnosis to Therapy'', in part by the National Science Centre, Poland (Chist-Era IV) under grant 2022/04/Y/ST6/00001, and in part by the National Science Centre, Poland under grant 2025/57/N/HS1/00480. Liesbeth Allein is further supported by a junior postdoctoral fellowship from the FWO under grant 12AGW26N.

\appendix

\section{Experimental Setup \& Reproducibility} \label{app:reproducibility} 

\begin{table}[h!]
    \centering
    \footnotesize
    \begin{tabular}{p{3cm}p{3.5cm}}
    \toprule
    \textbf{Hyperparameter} & \textbf{Value} \\
    \midrule
    \multicolumn{2}{c}{\textit{Ethos classification}} \\
    \midrule
       Model name & RoBERTa Large \\
       Model URL & \url{https://huggingface.co/FacebookAI/roberta-large} \\
       \# Layers & 24 \\ 
       Hidden size  & 1024 \\
       \# Attention heads & 16 \\
       \# Model parameters & 355M \\
       Loss function & Cross-entropy loss \\
       Optimizer & Adam \\
       Learning rate & 2e-5 \\
       \# Epochs & 6 \\
       Batch size & 32\\
    \midrule
    \multicolumn{2}{c}{\textit{Pathos classification}} \\
    \midrule
       Model name & RoBERTa Base \\
       Model URL & \url{https://huggingface.co/FacebookAI/roberta-base}\\ 
       \# Layers & 12 \\ 
       Hidden size  & 768 \\
       \# Attention heads & 12 \\
       \# Model parameters & 125M \\
       Loss function & Cross-entropy loss \\
       Optimizer & Adam \\
       Learning rate & 4e-5 \\
       \# Epochs & 5 \\
       Batch size & 32\\
    \bottomrule
    \end{tabular}
    \caption{Hyperparameter settings for finetuning the ethos and pathos classifiers.}
    \label{tab:hyperparameters}
\end{table}

\paragraph{Ethos and pathos classification} Table \ref{tab:hyperparameters} includes all model and finetuning hyperparameters for reproducing the ethos and pathos classification.  PolarIs dataset\footnote{https://osf.io/xjbaw/} is a collection of Reddit and Twitter discussions, centered around three topics: COVID-19 vaccines, climate change, and the 2016 U.S. elections. These social media posts were then split into individual sentences using the spaCy library. In total, it comprises 15,588 sentences annotated with appeals to ethos and pathos. Pre-processing involved the transformation of Twitter usernames into a common ``@user'' tag. Since the data is highly imbalanced (80\% of sentences are $et_s=0$, and 69\% are $pa_s=0$), we undersample the neutral category to constitute 50\% of labels for both ethos and pathos. For the purpose of this study, we sample 60\% of the data for training, 20\% for testing, and 20\% for validation in a label-stratified manner. 
All experiments, including finetuning of the RoBERTa models for ethos and pathos classification, were run on a NVIDIA Tesla T4 GPU available in Google Colab. 

\paragraph{Interpretations} All analyses were conducted over the entire OrigamIM dataset, for which we combined the train, validation, and test set provided by its creators. 

\paragraph{Evaluation and analysis}
All analyses were conducted in the Python programming language, and specifically the Sklearn library for calculating classification performance metrics\footnote{https://scikit-learn.org/stable/}, Statsmodels for executing the linear regression model\footnote{https://www.statsmodels.org/stable/index.html}, Scipy for statistical testing\footnote{https://scipy.org/}, and Transformers for model fine-tuning\footnote{https://pypi.org/project/transformers/}. 

\paragraph{Regression analyses with Gemini classification}
Below we present the results of two regression analyses with an alternative classification approach to ethos and pathos appeals, i.e. with the Gemini model (Tables \ref{tab:regression_variability_gemini} and \ref{tab:regression_gemini}). 

\begin{table}[h]
\centering
\setlength{\tabcolsep}{3pt} 
\begin{tabular}{lrrrr}
\hline
\textbf{Predictor} & \textbf{$\beta$} & \textbf{SE} & \textbf{$t$} & \textbf{$p$} \\
\hline
Intercept & 1.54 & 0.02 &  79.33 & *** \\
\textcolor{pathosorange}{[Pathos: +]} & 0.21 & 0.05 & 4.17 & *** \\ 
\textcolor{pathosorange}{[Pathos: -]} 
& 0.06 & 0.03 & 2.29 & **  \\
\hline

Intercept & 1.33 & 0.02 & 86.91 & *** \\
\textcolor{ethosblue}{[Ethos: +]} & 0.46 & 0.04 & 11.93 &  *** \\
\textcolor{ethosblue}{[Ethos: -]} & 0.37 & 0.03 & 13.98 & *** \\
\hline
\end{tabular} 
\caption{OLS regression predicting \textbf{variability in interpretation labels from sentence labels} with Gemini. Neutral labels are set as baselines. Statistical significance: *** ($p<.001$), ** ($p<.01$).}
\label{tab:regression_variability_gemini}
\end{table}

\begin{table}[h]
\centering
\setlength{\tabcolsep}{3pt} 
\begin{tabular}{lrrrr}
\hline
\textbf{Predictor} & \textbf{$\beta$} & \textbf{SE} & \textbf{$t$} & \textbf{$p$} \\
\hline
Intercept & 3.03 & 0.02 & 193.94 & *** \\
\textcolor{pathosorange}{[Pathos: +]} & 0.37 & 0.04 & 9.03 & ***  \\
\textcolor{pathosorange}{[Pathos: -]} & -0.14 & 0.02 & -6.57& *** \\ \hline

Intercept & 3.00 & 0.01 & 225.43 & *** \\
\textcolor{ethosblue}{[Ethos: +]} &  0.14 & 0.03 & 4.13 &  *** \\
\textcolor{ethosblue}{[Ethos: -]} & -0.10 & 0.02 & -4.29 & *** \\
\hline
\end{tabular} 
\caption{{OLS regression predicting \textbf{audience attitudes from sentence-level pathos and ethos appeals} with Gemini. Neutral labels are set as baselines. Statistical significance: *** ($p<.001$).}}
\label{tab:regression_gemini}
\end{table}

\section{Prompting Templates} \label{app:prompting_templates}
Prompt templates are given on an example of ethos annotation; templates for pathos are formulated in an analogous manner.

\paragraph{Zero-shot.}
Your task is to identify ethos appeals in a list of sentences. Ethos appeals are references to a speaker’s character or credibility.
These references may be:
  • favorable (supporting someone's credibility or character),
  • unfavorable (attacking or undermining credibility or character),
  • non-ethos (no reference to character/credibility).

\paragraph{Few-shot.}
Your task is to identify ethos appeals in a list of sentences. Ethos appeals are references to a speaker’s character or credibility.
These references may be:
  • favorable (supporting someone's credibility or character),
  • unfavorable (attacking or undermining credibility or character),
  • non-ethos (no reference to character/credibility).
Example of a favorable ethos appeal: Great initiative by MRIs. 
Example of a unfavorable ethos appeal: To me it's just a control/manipulation tactic by the government and elites. 
Example of a non-ethos appeal: A few months ago we had the worst flood our area has had in over 100 years.

\paragraph{Chain-of-thought.}
Your task is to identify ethos appeals in a list of sentences. Ethos appeals are references to a speaker’s character or credibility.
These references may be: 
  • favorable (supporting someone's credibility or character)
  • unfavorable (attacking or undermining credibility or character)
  • non-ethos (no reference to character/credibility)
Use step-by-step reasoning internally to identify ethos appeals and output the required labels. Below are three worked examples that illustrate the reasoning style. 

\noindent
\#Example 1 (unfavorable ethos) 
Sentence: "To me it's just a control/manipulation tactic by the government and elites." 

\noindent 
Demonstrated CoT: 
The sentence attacks the credibility and intentions of “the government and elites.”  → This is an unfavorable ethos appeal. 
Final output format: 
ID 1: unfavorable. 

\noindent
\#Example 2 (favorable ethos) 
Sentence: "Great initiative by MRIs."

\noindent
Demonstrated CoT: 
The sentence praises MRIs, implying they are doing something commendable.  
→ This supports their credibility/character. 
Final output format: 
ID 2: favorable.

\noindent
\#Example 3 (non-ethos) 
Sentence: "A few months ago we had the worst flood our area has had in over 100 years."

\noindent
Demonstrated CoT: 
The sentence reports an event, with no reference to character or credibility.  
→ This is non-ethos. 
Final output format: 
ID 3: non-ethos.

\noindent
Now process the following sentences using the same rules and output labels in the specified format.

\section{Manual Validation: Annotation Guidelines} \label{app:annotation_guidelines_validation}

{
The corpora are annotated at the sentence level for two rhetorical strategies from Aristotelian rhetoric: ethos (credibility appeals) and pathos (emotional appeals). The following guidelines come from the PolarIs dataset \citep{gajewska2024ethos}, which we use also for our manual verification of classification reliability in the target OrigamIM dataset. Here, the annotation is conducted on a randomly sampled 100 cases from the OrigamIM dataset and annotated independently by two trained raters. Both annotators are females with a previous experience in annotation studies and linguistic research, who voluntarily performed the task. 
\paragraph{Ethos Annotation.} The goal of ethos annotation is to identify sentences where a speaker references the credibility or character of an entity, either in a supportive or attacking way. The procedure for identifying such references is as follows. First, the rater should detect an ethos appeal if a sentence mentions an entity (a person/group/organization) and this entity is described in a favorable (support) or unfavorable (attack) manner. Second, the rater should mark the sentence as a support if the entity is mentioned in a favorable manner or as an attack if the entity is mentioned in an unfavorable manner. 
\paragraph{Pathos Annotation.} The goal of pathos annotation is to capture rhetorical appeals that aim to induce emotions in the hearer rather than simply express a sentiment. The procedure for identifying such appeals is as follows. First, the rater should determine whether the speaker’s language intends rhetorical gain through emotional cues. Second, the rater should mark positive pathos when the speaker intended to induce positive emotional states in the hearers or negative pathos when the speaker intended to induce negative emotional states in the hearers. Linguistic signals for the identification fo pathos appeals include emotion-eliciting words, phrases, and rhetorical figures. }

\section{Access to Existing and New Artifacts}

Our analyses relied on the OrigamIM dataset \citep{allein2024origamim}, which is licensed under the CC BY-SA 4.0 license. We used OrigamIM for research purposes only, which is consistent with its intended use. We retrieved the entire dataset from a publicly accessible GitHub repository\footnote{\url{https://github.com/laallein/origamIM}}. We will release the OrigamIM dataset with the labels for ethos and pathos inferred in this work with the same CC BY-SA 4.0 license in a public repository, for which we have oral consent from the creators of OrigamIM.

\section{Definitions of Variability Factors} \label{app:definition_clarification}
{Below we provide definitions of eight factors driving the variability between the sentence and interpretations beyond ethos and pathos labels.} 

{\begin{itemize}
    \item[] \textbf{Author opinion}: Sentences written in the first person are frequently rephrased into third-person statements, altering perceived speaker commitment and ethical stance reflected in the interpretations.
    \item[] \textbf{Value system differences}: Interpreters impose their own moral or ideological priorities onto the sentence, leading the same content to be framed as either ethically responsible or ethically problematic.
    \item[] \textbf{Political partisanship}:  References to political actors (e.g., Donald Trump) are often interpreted through partisan lenses, resulting in polarity shifts such as praise being rephrased into criticism.
    \item[] \textbf{Social identity or social position}: Mentions of race, gender or group membership activate different social meanings, affecting both perceived intent and emotional appeal.
    \item[] \textbf{Out-of-context effects}: Sentences that are very short or lack the necessary contextual information to be understood on their own lead to interpreters having to infer the referent, stance or target of the sentence. This, in turn, may lead to divergent or incorrect assumptions about what or who the sentence is addressing.
    \item[] \textbf{Rhetorical questions}: Rhetorical questions are frequently normalized into declarative claims, changing their persuasive force and associated ethos or pathos labels.
    \item[] \textbf{Personal experience}: This is closely related to author opinion. However, personal experience specifically refers to cases where sentence authors invoke anecdotal or lived experiences to support a broader claim. While author opinion reflects a general personal stance, personal experience grounds that stance in a specific situation the author reports having encountered. Interpretations of such sentences vary depending on whether readers treat the anecdote as credible evidence, a subjective account, or a rhetorical device, which can lead to differences in how ethos and pathos are assigned.
    \item[] \textbf{Sarcasm}: ironic or mocking intent is not consistently detected, resulting in marked disagreement in emotional and ethical labeling across interpretations. 
    \end{itemize}
}

\section{On the Use of AI Assistants in Research or Writing}

AI assistants were used in this research to assist in writing (ChatGPT, Copilot). After using these tools/services, the authors reviewed and edited the content as needed and take full responsibility for the content of the publication.

\end{document}